\documentclass[draftclsnofoot,peerreview,11pt,onecolumn,doublespace]{IEEEtran}
\usepackage{subfig}
\usepackage{amssymb}
\usepackage{cite}
\usepackage[pdftex]{graphicx}
\graphicspath{{./}{figures/}}
\usepackage[table]{xcolor}
\usepackage{arydshln}
\usepackage{amsmath}
\usepackage{amssymb}
\usepackage{url}
\usepackage{setspace}
\usepackage{fixltx2e}
\usepackage{rotating}
\usepackage{multirow}
\usepackage{floatrow}
\usepackage[font=small,labelfont=bf]{caption}
\newfloatcommand{capbtabbox}{table}[][\FBwidth]
\DeclareMathOperator*{\argmin}{arg\,min}
\newcolumntype{M}{>{$\vcenter\bgroup\hbox\bgroup}c<{\egroup\egroup$}}

\usepackage{blindtext}
\usepackage{lscape}

\def\ie{\emph{i.e. }}

\def\SURE{\textrm{SURE}}
\def\PSURE{\textrm{PSURE}}
\def\BSURE{\textrm{BSURE}}

\newcommand{\im}[1]{\mathbf{#1}}
\newcommand{\est}[1]{\widehat{#1}}

\usepackage{algorithm}
\usepackage{algorithmic}
\usepackage{caption}
\begin{document}
\setlength{\intextsep}{3pt} 
\setlength{\textfloatsep}{3pt}
\setlength{\abovecaptionskip}{3pt}
\setlength{\belowcaptionskip}{3pt}
\setlength{\dbltextfloatsep}{3pt}
\title{Blockwise SURE Shrinkage for Non-Local Means}
\author{Yue~Wu, {Brian~Tracey}, {Premkumar Natarajan}
        and Joseph~P.~Noonan
\thanks{ Brian Tracey and Joseph P. Noonan are with the department
of electrical and computer engineering, Tufts university, 161 College Ave,
Medford, MA 02155; e-mail: ywu03@ece.tufts.edu. Premkumar Natarajan is
with the Raytheon BBN technologies, 10 Moulton St., Cambridge, MA 02138. Yue Wu was with the department of electrical and computer engineering at Tufts university, but is now with the Raytheon BBN technologies.} }
\maketitle

\begin{abstract}
In this letter, we investigate the shrinkage problem for the non-local means (NLM) image denoising. In particular, we derive the closed-form of the optimal blockwise shrinkage for NLM that minimizes the Stein's unbiased risk estimator (SURE). We also propose a constant complexity algorithm allowing fast blockwise shrinkage. Simulation results show that the proposed blockwise shrinkage method improves NLM performance in attaining higher peak signal noise ratio (PSNR) and structural similarity index (SSIM), and makes NLM more robust against parameter changes. Similar ideas can be applicable to other patchwise image denoising techniques such as \cite{multipatch} and \cite{dabov2006image}.
\end{abstract}

\IEEEpeerreviewmaketitle

\section{Introduction}
In recent years, the most popular image denoising algorithms include the NLM method \cite{NLM0,NLM1} and the BM3D method \cite{dabov2006image,dabov2007image}. Both utilize adaptive patch weighting and aggregations during denoising, while NLM denoising  is pixelwise whereas BM3D is patchwise. Although satisfactory denoising results using both methods are widely reported, determining an appropriate set of parameters is truly a nontrivial task: 1) improper parameter choices might make these methods perform badly, and 2) the entire parameter space is huge to be fully explored. The importance of making NLM and BM3D more robust and less sensitive to parameter changes are thus self-exploratory.

Many efforts of selecting parameters automatically or improving method robustness have been discussed recently, especially for the NLM. \cite{SURE} derived the closed-form of NLM-SURE and used the SURE to pick the proper bandwidth parameter. Although \cite{SURE} gave an empirical choice of the bandwidth parameter, the optimal choice (also dependent on image content) still requires looping over NLM for many times. \cite{multipatch} introduced the multi-patch NLM to overcome the denoising artifacts by combining results of using various patch shapes and sizes. However, this work also requires denoising an image many times, once for each patch shape. \cite{JSCPW} introduced the James-Stein type NLM shrinkage and made the NLM less sensitive to the bandwidth parameter changes. Although this earlier work of ours requires one-time processing only, it is a sub-optimal solution, because its shrinkage is made with respect to one of the SURE terms but not all of them.

In this paper, we propose a new SURE-based NLM shrinkage technique. It only requires one-time NLM denoising and SURE computations, but shrinks an image optimally with respect to the denoising risk. It enhances  NLM performance and improves the NLM robustness against parameter changes. The rest of the letter is organized as follows: Sec. II briefly reviews the classic NLM and the NLM-SURE; Sec. III derives the closed-form of optimal blockwise shrinkage, proposes the SURE-based pixel aggregations, and discusses the fast implementation; Sec. IV shows our simulation results; and we conclude the letter in Sec. V.

\section{Preliminary}
Assume a clean image $\im{x}\!=\!\{\!x_l\!\}_{l\in\mathbb I}$ is contaminated by i.i.d. zero-mean Gaussian noises with an unknown variance, namely
\begin{equation}\label{eq.gaussiannoise}
    y_l = x_l+n_l, \textrm{\, and \,} n_l\sim{\cal{N}}(0,\sigma^2)
\end{equation}
where $\im{y}\!=\!\{\!y_l\!\}_{l\in\mathbb I}$ is noisy observation of $\im{x}$ and $n_l$ is the noise on $l$th pixel. The classic NLM \cite{NLM0,NLM1} estimates $\est{x}_l$ as the weighted sum of $y_l$'s noisy neighbors within a prescribed search region $\mathbb{S}$ (typically a square or a rectangle), \ie
\begin{equation}\label{eq.nlm}
\est{x_l} = \textstyle\sum_{k\in\mathbb{S}}{w_{l,k}y_{k}}/(\textstyle{\sum_{k\in\mathbb{S}}w_{l,k}})
\end{equation}
where each weight is computed by quantifying the similarity between two local patches around noisy pixels $y_l$ and $y_k$ as shown in \eqref{eq.nlmweight} with the local patch parameter $\mathbb{P}$, and the bandwidth parameter $h$.
\begin{equation}\label{eq.nlmweight}
    w_{l,k} = \exp\big(-{\textstyle\sum_{j\in\mathbb{P}}({y_{l+j}-y_{k+j})^2}/ 2h}\big)
\end{equation}

The NLM-SURE estimator \cite{SURE} is a powerful tool predicting the denoising risk without knowing the clean image. Since it can be obtained along with the computation of $\est{\im{x}}$, it has been extensively used in parameter selections. The NLM-SURE is of the closed-form \eqref{eq.SURE}
\begin{equation}\label{eq.SURE}
    \textrm{SURE}(\est{\mathbf x}) = \|{\mathbf y}-\est{\mathbf x}\|_2^2/|{\mathbb I}|+2\sigma^2\textrm{div}_{\mathbf y}\{\est{\mathbf x}\}/|{\mathbb I}|-\sigma^2
\end{equation}
where $|\cdot|$ denotes the cardinality of a set, and
\begin{equation}
\textrm{div}_{\mathbf y}\{\est{\mathbf x}\}\!\!=\!\!\textstyle\sum_{l\in \mathbb{I}}{\partial \est{x_l}/ \partial y_l}
\end{equation} is the divergence term with each ${\partial \est{x_l}/\partial y_l}$ defined in \eqref{Eq.partialDIV}
\begin{equation}\label{Eq.partialDIV}
    {\partial \est{x_l}\over \partial y_l} = {{\est{x_l^2}}-\est{x_l}^2\over h}+{w_{l,l}\over W_l}+\sum_{i\in \mathbb{P}} {w_{l,l-i}\over W_lh}(y_l-y_{l+i})(\est{x}_l-y_{l-i})
\end{equation}
with $W_l$ as the summed weights and ${\est{x_l^2}}$ as the $2$nd moment.
\begin{equation}\label{eq.weightSum}
    W_l = \textstyle\sum_{k\in\mathbb{S}_l} w_{l,k} \,\textrm{ and }\,\est{x_l^2} = {\textstyle\sum_{k\in \mathbb{S}_l} w_{l,k}y_k^2}/{W_l}
\end{equation}

\section{Blockwise SURE Shrinkage}
Assume we have performed NLM denoising with NLM-SURE, obtaining an initial denoised image $\im{\est{x}}\!=\!\{\est{x_l}\}_{l\in\mathbb{I}}$ with its denoising risk SURE$(\im{\est{x}})$ including all intermediate terms in \eqref{eq.nlm} and \eqref{eq.SURE}. Notice that SURE$(\im{\est{x}})$ is actually estimated as the average of all pixelwise SURE (PSURE) map in \eqref{eq.psure}:
\begin{equation}\label{eq.SUREinp}
    \SURE(\im{\est{x}}) = \textstyle\sum_{l\in{\mathbb{I}}}\PSURE(\est{x_l})/|\mathbb{I}|.
\end{equation}
\begin{equation}\label{eq.psure}
    \PSURE(\est{x_l}) = ({{y_l}}-\est{x_l})^2+2\sigma^2{\partial \est{x_l}/ \partial y_l}-\sigma^2.
\end{equation}
We are interested in forming a better estimation of the clean image by shrinking the initial denoised image towards the noisy observation. Mathematically, this shrinkage process on the pixel-level can be written as
\begin{equation}\label{eq.xlprime}
    \est{x}_l' = (1-q_l)\est{x_l}+q_l y_l
\end{equation}
where $q_l$ is the shrinkage parameter. Since we know the risks of $\est{x_l}$ and $y_l$ (they are PSURE(${\est{x_l}}$) and $\sigma^2$, respectively) on the right side of shrinkage estimator \eqref{eq.xlprime}, it is natural to ask what is PSURE($\est{x}'_l$), the risk after shrinkage with parameter $q_l$.
Once we know the answer to this important question, we may find an optimal $q_l$ that minimizes PSURE($\est{x}'_l$). In the rest of this section, we shall propose our answer to this question and show how to achieve optimal blockwise shrinkage.

\subsection{Optimal SURE Shrinkage}
To find PSURE($\est{x}'_l$), first notice the PSURE after shrinkage is of the form
\begin{eqnarray}\label{eq.psureprim}
    \PSURE(\est{x}'_l) \!\!&\!\!=\!\!&\!\! ({{y_l}}-\est{x}'_l)^2+2\sigma^2{\partial \est{x}'_l/ \partial y_l}-\sigma^2
\end{eqnarray}
whose terms $({{y_l}}-\est{x}'_l)^2$ and ${\partial \est{x}'_l/ \partial y_l}$ can be found as
\begin{eqnarray}
  ({{y_l}}-\est{x}'_l)^2 \!\!&\!\!=\!\!&\!\!  (1-q_l)^2({{y_l}}-\est{x_l})^2 \\
  {\partial \est{x}'_l/\partial y_l} \!\!&\!\!=\!\!&\!\!  (1-q_l){\partial \est{x_l}/ \partial y_l}+q_l
\end{eqnarray}
by simply substituting \eqref{eq.xlprime}. After simplifications, we obtain $\PSURE(\est{x}'_l)$ as a function of the shrinkage parameter, \ie
\begin{equation}\label{eq.fql2}
    \PSURE(\est{x}'_l|q_l) = a_{l,2}q_l^2+2a_{l,1}q_l+a_{l,0}
\end{equation}
with known coefficients
\begin{eqnarray}
  a_{l,2} \!\!&\!\!=\!\!&\!\! (y_l-\est{x}_l)^2\\
  a_{l,1} \!\!&\!\!=\!\!&\!\! \sigma^2{\partial \est{x_l}/ \partial y_l}-\PSURE(\est{x_l})\\
  a_{l,0} \!\!&\!\!=\!\!&\!\! \PSURE(\est{x_l}).
\end{eqnarray}
Since $\PSURE(\est{x}'_l|q_l)$ is quadratic and concave up ($a_{l,2}\geq 0$), there exists a unique global minimum at $q_l^*$ where
\begin{equation}\label{eq.qlstar}
    q_l^* = \textstyle \argmin_{q_l} \PSURE(\est{x}'_l|q_l) = -a_{l,1}/a_{l,2}.
\end{equation}
Although in theory $q_l^*$ provides the optimal shrinkage parameter, it requires an accurate $\PSURE(\est{x_l})$ estimation, which might deviate far away from its true value in practice. Fortunately, according to the unbiased estimator nature of SURE, the more pixels we have in an image region, the more accurate risk estimation we achieve. We thus define the blockwise SURE (BSURE) over an image block $\im{b}_l^x = \{\est{x}_{l+j}|j\in\mathbb{B}_l\}$ as
\begin{eqnarray}\label{eq.bsure}
    \BSURE(\im{b}_l^x)\!=\!\textstyle \textstyle\sum_{j\in{\mathbb{B}_l}}\PSURE(\est{x}_{l+j})/|\mathbb{B}_l|.
\end{eqnarray}
Assume we uniformly shrink the pixels in $\im{b}_l$ with respect to a parameter $p_l$, \ie $\forall j\in\mathbb{B}_l, \exists q_{l+j} = p_l$. Then the risk of using new denoised block after shrinkage using $p_l$ is\newline\vspace{-11pt}
\begin{eqnarray}\label{eq.bsure}
   \nonumber \BSURE(\im{b}_l^{x'}|p_l)\!\!&\!\!=\!\!&\!\! \textstyle \textstyle\sum_{j\in{\mathbb{B}_l}}\PSURE(\est{x}_{l+j}'|p_l)/|\mathbb{B}_l|\\   \!\!&\!\!=\!\!&\!\! \textstyle (A_{l,2}p^2_l\!+\!2A_{l,1}p_l\!+\!A_{l,0})/|\mathbb{B}_l|
\end{eqnarray}
with coefficient terms
\begin{eqnarray}
  A_{l,2}\!\!&\!\!=\!\!&\!\!\textstyle \sum_{j\in{\mathbb{B}_l}}a_{l+j,2} \\
  A_{l,1} \!\!&\!\!=\!\!&\!\!\textstyle \sum_{j\in{\mathbb{B}_l}}a_{l+j,1} \\
  A_{l,0} \!\!&\!\!=\!\!&\!\!\textstyle \sum_{j\in{\mathbb{B}_l}}a_{l+j,0}.
\end{eqnarray}
Again, the BSURE is a function of the shrinkage parameter $p_l$, and BSURE is minimized when
\begin{equation}\label{eq.plstar}
    p_l^* =\textstyle \argmin_{p_l} \BSURE(\im{b}_l^{x'}|p_l) = -A_{l,1}/A_{l,2}
\end{equation}
which is the optimal shrinkage parameter for image block $\im{b}_l$. We denote the new block after optimal shrinkage as
\begin{equation}\label{eq.bsureblx}
    \im{b}^{x^*}_l = (1-p_l^*)\im{b}_l^x+p_l^*\im{b}_l^y.
\end{equation}

\subsection{SURE-Based Pixel Aggregations}
When the entire image $\est{\im{x}}$ is considered as one block, then Eq. \eqref{eq.plstar} directly gives the optimal solution for global shrinkage. However image information is contained within local pixel blocks, so it is more plausible to shrink image blocks locally. Yet the derived optimal blockwise shrinkage (BSS) is applicable to an arbitrary local block, we follow the BM3D fashion and use overlapping local blocks. Consequently, these overlapping blocks lead to an overcomplete problem in determining the final estimated pixels, because each initially denoised pixel $\est{x}_k$ might be in multiple blocks while each block gives one candidate $\est{x}_k'$ via \eqref{eq.bsureblx}. This overcomplete problem can be approached by reestimating the final denoised pixels $\est{x}_k''$s from all BSS $\est{x}_k'$s \cite{dabov2006image}. To simplify discussion, we pretend $\im{b}^{x^*}_l$ is an image of the same size as the noisy image, but with all zeros for those pixels outside of $\mathbb{B}_l$, \ie
\begin{equation}\label{eq.}
    \est{x}_{k|\mathbb{B}_l}'={b}^{x^*}_l[k] = \left\{
    \begin{array}{r}
    0,\,\textrm{if}\, k-l\notin\mathbb{B}_l\\
    (1-p^*_l)\est{x}_k+p^*_ly_k,\textrm{otherwise}
    \end{array}
    \right.
\end{equation}
Let $\mathbb{R}_k\!=\!\{l|k-l\in\!\mathbb{B}_l\!\}$ be the index set of all blocks containing the pixel $\est{x}_k$. We then reestimate ${\est{x}}''$ as the weighted average of all BSS $\est{x}_k'$s
\begin{eqnarray}\label{eq.agg}
    \est{x}_k''\!\!&\!\!=\!\!&\!\!\textstyle\left(\sum_{l\in\mathbb{R}_k}v_l \cdot\est{x}_{k|\mathbb{B}_l}'\right)/(\sum_{l\in\mathbb{R}_k}v_l)
    \end{eqnarray}
where the aggregation weight for each block is computed from
\begin{equation}\label{eq.vl}
    v_l = \exp(-\BSURE(\im{b}^{x^*}_l)/\sigma^2).
\end{equation}
In this way, we make the SURE-based pixel aggregations and obtain the final denoised image $\im{\est{x}}''$.
\vspace{-4pt}
\subsection{Implementation}
With regards to implementation,  it is desired to have 1) fast BSS computations, and 2) straightforward parameter selection $\mathbb{B}_l$s. To achieve both goals, we use all square size blocks $\mathbb{B}_l$, and repeat BSS process with growing blocks until the shrinkage converges. A pseudo code of the described implementation is given in Algorithm 1. Specifically speaking, the integral image (II) \cite{II,IINLM} (line 3) is a fast algorithm for computing arbitrary rectangular sums. In particular, it requires 2 operations/pixel to construct an II and 3 operations/pixel to extract the sum of pixels within a rectangular region. The SURE-based aggregation \eqref{eq.agg} can be done sequentially (line 12) because its equivalent form is
\begin{eqnarray}\label{}
    \est{x}_k''\!\!&\!\!=\!\!&\!\! \est{x}_k+{\sum_{l\in\mathbb{R}_k}\!v_l p^*_l\over\sum_{l\in\mathbb{R}_k}v_l}(y_k\!-\!\est{x}_k).
    \end{eqnarray}

\begin{algorithm}[H]
\caption{Fast Blockwise SURE Shrinkage}
\scriptsize\centering
\begin{algorithmic}[1]
\REQUIRE $\PSURE(\est{x_l})$ map, divergence ${\partial \est{x_l}/ \partial y_l}$, initial result $\est{\im{x}}$, noisy image $\im{y}$, tolerance $\delta$ and image size $M$.
\ENSURE  blockwise shrinkage image $\est{\im{x}}''$
\STATE initialize $blkSize=7$; $\est{\im{x}}''=\im{0}$; $\im{V}=\im{0}$; $\im{S}=\im{0}$; $\im{t}=\im{\est{x}}$.
\STATE for all $l\in\mathbb{I}$, compute $a_{l,2}$ $a_{l,1}$ and $a_{l,0}$ using (17) (18), and (19), respectively.
\STATE construct integral images $\im{\rm II}(a_{l,2})$, $\im{\rm II}(a_{l,1})$, and $\im{\rm II}(a_{l,0})$
\WHILE{}
\STATE for $n\in\{0,1,2\}$, compute coefficients $A_{l,n}$ from $\im{\rm II}(a_{l,n})$.
\STATE compute optimal shrinkage $p^*_l$ using \eqref{eq.plstar} for square block of size $blksize$.
\STATE compute $\BSURE(\im{b}_l^{x'}|p_l^*)$ using \eqref{eq.bsure}.
\STATE for all $l\in\mathbb{I}$, compute its weights of aggregation $v_l$ using \eqref{eq.vl}.
\STATE compute the weight sum $vv_k\leftarrow\sum_{l\in\mathbb{R}_k}v_l$.
\STATE update the accumulated weight ${V}_k\leftarrow{V}_k+{vv}_k$.
\STATE compute the shrinkage sum $ss_k \leftarrow\sum_{l\in\mathbb{R}_k}v_lp_l^*$
\STATE update the accumulated shrinkage sum ${S}_k\leftarrow{S}_k+{ss}_k$.
\STATE update $\est{{x}}''_k\leftarrow \est{x}_k+(y_k-\est{x}_k)S_k/V_k$ using \eqref{eq.agg}
\IF{$\sum_{l\in\mathbb{I}}(\est{x}''_l-t_l)^2/|\mathbb{I}|\leq\delta$ \&\& $blkSize\leq M$}
    \RETURN $\im{\est{{x}}}''$
\ELSE
\STATE { $blkSize\leftarrow blkSize+1$, $t_l\leftarrow {\est{{x}}}''_l$.}
\ENDIF
\ENDWHILE
\end{algorithmic}
\end{algorithm}
\noindent The convergence condition is set to $10^{-4}$ in experiments. This algorithm usually takes about 4-20 rounds to converge. The complexity of each round is approximately 50 operations/pixel (to be precise, the arithmetic complexity of line 5 to line 18 are 15, 2, 8, 4, 5, 1, 6, 1, 3, 2, and 2, respectively). Because integral images are used, the complexity of each shrinkage round is independent of the used block size. For a 256$\times$256 grayscale image, the time complexity of executing the fast NLM \cite{IINLM}, the SURE computation, and the additional BSS/round (from 300 realizations with 2.4GHz cpu) are .4309$\pm$.0077, .0181$\pm$.0008, and .0177$\pm$.0016 seconds, respectively.

\section{Experiments}
All following simulations are done under the MATLAB r2010b environment. The MATLAB BSS implementation can be provided upon request. Because the dual problem of the center pixel weight and the shrinkage estimation in NLM \cite{JSCPW}, we compare the classic and recent NLM CPW solutions including the standard (std) CPW \cite{NLM0,NLM1}, the zero CPW \cite{CPWstein}, the max CPW \cite{CPWstein}, the heuristic (heru) CPW \cite{JSCPW}, the stein CPW \cite{CPWstein} and the local James-Stein (ljs) shrinkage \cite{JSCPW}. Technically speaking, all these test methods differ from each other only in the weights of using noisy pixels (\ie different $q_l$s in \eqref{eq.xlprime}).  We test denoising performance of each method by using simulated noisy images with the noise level, \ie standard deviation $\sigma$, ranging from 10 to 60 under various NLM parameter combinations. Specifically, the NLM patches $\mathbb{P}$ used in simulations vary from 3$\times$3 to 7$\times$7, the search region is fixed at 15$\times$15, and the bandwidth parameter $h$ is chosen from 5\% to 200\% of $|\mathbb{P}|\sigma^2$. The quality of each denoised image is evaluated by using the PSNR \cite{SURE} and the SSIM \cite{ssim}. The best scores over all $h$s for each method under different parameter combinations are given in Table I (methods with best scores are underlined).
\footnotetext[1]{Test images are available at \url{http://www.cs.tut.fi/~foi/GCF-BM3D/BM3D_images.zip} as the date of March/22/2013. }

This table shows two general trends: the proposed BSS method 1) attains the best overall performance in PSNR/SSIM scores; and 2) are more robust against the patch size change than other methods. The proposed BSS method put additional .3 to 1.1 dB on the best PSNR scores and 2\% to 8\% on the best SSIM score of using the standard NLM CPW. It is worthwhile to point out these gains on the best standard NLM scores are not trivial, and to some extend these BSS scores with simple shrinkage estimations are comparable to or better than more complicated NLM variants, for examples the linear expansion with six NLMs (see Table II in \cite{van2011nonlocal}), and the multi-patch NLMs (see Table 5 in \cite{multipatch}), both of which requires multi-rounds of NLM denoising.

Fig. 1 shows the method sensitivity to the patch size parameter, where each method data point is averaged from the sensitivity scores for all six images, with each method image sensitivity score is the standard deviation of scores in Table I for all three patch sizes of the corresponding method. It is obvious that the average standard deviation of PSNR and SSIM scores of proposed BSS method is much smaller than others, and also less linearly dependent on the noise level. 

Sample denoised images and the corresponding difference images from the clean images are given in Fig. 2. It is noticeable that the BSS result is sharper on edges (see \textit{lenna}'s hair and \textit{Hello World}), while smoother on homogeneous regions (see \textit{sphere}). Similar results are also observed of using other images and NLM parameters. This shows the BSS method makes NLM more robust against different image contents.
\section{Conclusion}
In this letter, we derived the analytic form of the optimal blockwsie SURE shrinkage for the NLM method. In this way, the optimal shrinkage parameters can be easily and efficiently computed from the SURE map of an initially denoised image, and allows a better estimation of the clean image without rerunning NLM denoising. Although in experiment we report the best scores by exhaustively searching the $h$ space, one may simply use the empirical optimal $h\!\approx\!|\mathbb{P}|\sigma^2/2$ in \cite{SURE} instead. Performance scores of using these empirical $h$ are close to reported ones. Consequently, one can take advantages brought by NLM-SURE and BSS, and eliminate both parameters $h$ and $\mathbb{P}$ in NLM without trading off overall performance.

Because SURE estimation improves with increasing block size, BSS performance can be further improved if disjoint but similar regions are identified and used for blockwise shrinkage.  Our initial attempts at using disjoint homogeneous regions show promise, especially for pixels near edges.  Further progress in this area requires a fast segmentation tool that gives robust disjoint partitions. The major difference between the BSS pixel aggregation \eqref{eq.agg} and that of BM3D \cite{dabov2006image,dabov2007image} is that our aggregations are made with respect to the SURE optimality instead of heuristics. This raises up an interesting question how to improve BM3D using the proposed BSS idea, and we shall explore this direction in future.

\begin{landscape}
\centering
\begin{table}[!htbp]
\tiny\centering
\caption{Performance comparisons for Non-local means with various CPW/shrinkage techniques}
\begin{tabular}{@{}M@{}|@{}m{0.035cm}@{}M@{}|@{}m{0.035cm}@{}M@{}m{0.035cm}@{}|@{}m{0.035cm}@{}M@{}m{0.035cm}@{}|@{}m{0.035cm}@{}M@{}m{0.035cm}@{}|@{}m{0.035cm}@{}M@{}m{0.035cm}@{}|@{}m{0.035cm}@{}M@{}m{0.035cm}@{}|@{}m{0.035cm}@{}M@{}m{0.035cm}@{}|@{}m{0.035cm}@{}M@{}m{0.035cm}@{}|@{}m{0.035cm}@{}M@{}m{0.035cm}@{}|@{}m{0.035cm}@{}M@{}m{0.035cm}@{}|@{}m{0.035cm}@{}M@{}m{0.035cm}@{}|@{}m{0.035cm}@{}M@{}m{0.035cm}@{}|@{}m{0.035cm}@{}M@{}m{0.035cm}@{}|@{}m{0.035cm}@{}M@{}m{0.035cm}@{}|@{}m{0.035cm}@{}M@{}m{0.035cm}@{}|@{}m{0.035cm}@{}M@{}m{0.035cm}@{}|@{}m{0.035cm}@{}M@{}m{0.035cm}@{}|@{}m{0.035cm}@{}M@{}m{0.035cm}@{}|@{}m{0.035cm}@{}M@{}m{0.035cm}@{}|
|@{}m{0.035cm}@{}M@{}m{0.035cm}@{}|@{}m{0.035cm}@{}M@{}m{0.035cm}@{}|@{}m{0.035cm}@{}M@{}m{0.035cm}@{}|@{}m{0.035cm}@{}M@{}m{0.035cm}@{}|@{}m{0.035cm}@{}M@{}m{0.035cm}@{}|@{}m{0.035cm}@{}M@{}m{0.035cm}@{}|@{}m{0.035cm}@{}M@{}m{0.035cm}@{}|@{}m{0.035cm}@{}M@{}m{0.035cm}@{}|@{}m{0.035cm}@{}M@{}m{0.035cm}@{}|@{}m{0.035cm}@{}M@{}m{0.035cm}@{}|@{}m{0.035cm}@{}M@{}m{0.035cm}@{}|@{}m{0.035cm}@{}M@{}m{0.035cm}@{}|@{}m{0.035cm}@{}M@{}m{0.035cm}@{}|@{}m{0.035cm}@{}M@{}m{0.035cm}@{}|@{}m{0.035cm}@{}M@{}m{0.035cm}@{}|@{}m{0.035cm}@{}M@{}m{0.035cm}@{}|@{}m{0.035cm}@{}M@{}m{0.035cm}@{}|@{}m{0.035cm}@{}M@{}m{0.035cm}@{}}
  \hline
  &&&&\multicolumn{53}{c||}{\bf Peak Signal Noise Ratio (dB)}&\multicolumn{52}{c}{\bf Structural Similarity Index (\%)}\\\hline
 &&&&\multicolumn{17}{c|}{\bf Patch 3$\times$3 }&\multicolumn{18}{c|}{\bf Patch 5$\times$5 }&\multicolumn{18}{c||}{\bf Patch 7$\times$7 }&\multicolumn{18}{c|}{\bf Patch 3$\times$3 }&\multicolumn{18}{c|}{\bf Patch 5$\times$5 }&\multicolumn{16}{c}{\bf Patch 7$\times$7 }\\ \hline &&$\sigma$&&\bf{10}&&&\bf{20}&&&\bf{30}&&&\bf{40}&&&\bf{50}&&&\bf{60}&&&\bf{10}&&&\bf{20}&&&\bf{30}&&&\bf{40}&&&\bf{50}&&&\bf{60}&&&\bf{10}&&&\bf{20}&&&\bf{30}&&&\bf{40}&&&\bf{50}&&&\bf{60}&&&\bf{10}&&&\bf{20}&&&\bf{30}&&&\bf{40}&&&\bf{50}&&&\bf{60}&&&\bf{10}&&&\bf{20}&&&\bf{30}&&&\bf{40}&&&\bf{50}&&&\bf{60}&&&\bf{10}&&&\bf{20}&&&\bf{30}&&&\bf{40}&&&\bf{50}&&&\bf{60}\\\hline
\multirow{7}{*}{\begin{sideways}{\bf\textit{cameraman}}\end{sideways}}&&\bf{zero}
&&	30.77	&&&	28.95	&&&	27.20	&&&	25.65	&&&	24.35	&&&	23.27	&&&	28.51	&&&	27.88	&&&	26.91	&&&	25.73	 &&&	24.59	&&&	23.60	&&&	27.92	&&&	27.37	&&&	26.56	&&&	25.59	&&&	 24.55	&&&	23.64	&&&	89.59	&&&	82.17	 &&&	76.66	&&&	71.48	&&&	66.22	&&&	61.17	&&&	87.52	&&&	82.17	&&&	77.48	&&&	72.76	&&&	68.38	&&&	64.39	 &&&	85.44	&&&	81.60	&&&	 77.69	&&&	73.42	&&&	69.12	&&&	65.15	\\
&&\bf{std}&&	33.35	&&&	29.57	&&&	27.39	&&&	25.70	&&&	24.33	&&&	23.20	&&&	32.96	&&&	29.21	&&&	27.38	&&&	25.85	 &&&	24.58	&&&	23.51	&&&	32.63	&&&	28.83	&&&	27.07	&&&	 25.65	&&&	24.45	&&&	23.42	&&&	90.91	&&&	82.50	 &&&	76.59	&&&	70.99	&&&	65.34	&&&	59.96	&&&	91.74	&&&	83.34	&&&	78.00	&&&	72.66	&&&	67.72	&&&	63.45	 &&&	91.48	&&&	 83.23	&&&	78.46	&&&	73.56	&&&	68.49	&&&	64.01	\\
&&\bf{max}&&	32.45	&&&	29.34	&&&	27.29	&&&	25.65	&&&	24.31	&&&	23.21	&&&	30.73	&&&	29.01	&&&	27.38	&&&	25.91	 &&&	24.63	&&&	23.58	&&&	30.32	&&&	28.77	&&&	27.33	&&&	 25.99	&&&	24.76	&&&	23.72	&&&	90.28	&&&	82.23	 &&&	76.51	&&&	71.02	&&&	65.45	&&&	60.11	&&&	89.46	&&&	82.72	&&&	77.52	&&&	72.49	&&&	67.87	&&&	63.71	 &&&	88.08	&&&	 82.75	&&&	78.06	&&&	73.35	&&&	68.73	&&&	64.58	\\
&&\bf{heru}&&	33.32	&&&	29.59	&&&	27.38	&&&	25.71	&&&	24.36	&&&	23.26	&&&	33.06	&&&	29.40	&&&	27.62	&&&	 26.09	&&&	24.76	&&&	23.67	&&&	32.77	&&&	29.04	&&&	27.45	&&&	 26.10	&&&	24.92	&&&	23.84	&&&	90.59	&&&	 82.45	&&&	76.65	&&&	71.24	&&&	65.75	&&&	60.52	&&&	91.47	&&&	83.29	&&&	78.06	&&&	72.87	&&&	68.22	&&&	 64.05	&&&\underline{91.60}	 &&&	83.38	&&&	78.78	&&&	74.06	&&&	69.30	&&&	65.06	\\
&&\bf{stein}&&	32.94	&&&	29.34	&&&	27.29	&&&	25.65	&&&	24.31	&&&	23.21	&&&	32.61	&&&	29.07	&&&	27.38	&&&	 25.91	&&&	24.63	&&&	23.58	&&&	32.36	&&&	28.97	&&&	27.33	&&&	 25.99	&&&	24.76	&&&	23.72	&&&	90.32	&&&	 82.23	&&&	76.51	&&&	71.02	&&&	65.45	&&&	60.11	&&&	90.43	&&&	82.72	&&&	77.52	&&&	72.49	&&&	67.87	&&&	 63.71	&&&	90.22	&&&	82.78	 &&&	78.06	&&&	73.35	&&&	68.73	&&&	64.58	\\
&&\bf{ljs}&&	33.35	&&&	29.67	&&&	27.64	&&&	25.97	&&&	24.89	&&&	23.33	&&&\underline{33.26}	&&&	29.55	&&&	27.55	&&&	 26.04	&&&	24.77	&&&	23.72	&&&	\underline{33.19}	 &&&\underline{29.48}	&&&	27.53	&&&	26.08	&&&	24.85	&&&	23.80	&&&	91.72	 &&&	\underline{84.49}	&&&	78.31	&&&	72.80	&&&	67.64	&&&	62.76	&&&	91.67	&&&\underline{84.48}	&&&	 78.21	&&&	73.74	&&&	 69.00	&&&	64.59	&&&	91.52	&&&\underline{84.33}	&&&	78.22	&&&	74.07	&&&	69.31	&&&	64.91	\\
&&\bf{bss}&&\underline{33.64}	&&&\underline{29.86}	&&&\underline{27.75}	&&&\underline{26.21}	&&&\underline{25.06}	&&&\underline{24.02}	&&&	33.22	 &&&\underline{29.63}	 &&&\underline{27.74}	&&&\underline{26.30}	&&&\underline{25.07}	&&&\underline{24.02}	&&&	32.92	&&&	29.33	&&&\underline{27.56}	 &&&\underline{26.21}	&&&\underline{25.04}	 &&&\underline{24.02}	&&&\underline{91.93}	&&&	84.45	&&&	\underline{78.52}	&&&\underline{73.51}	&&&\underline{68.36}	 &&&\underline{64.22}	&&&\underline{92.02}&&& 83.90	 &&&\underline{79.07}	&&&\underline{74.77}	&&&\underline{70.21}	&&&\underline{65.43}	&&&\underline{91.60}	 &&&	84.16	&&&\underline{78.87}	&&&\underline{74.77}	 &&&\underline{70.56}	&&&\underline{66.02}	\\
\hline
\multirow{7}{*}{\begin{sideways}{\bf\textit{montage}}\end{sideways}}&&\bf{zero}
&&	33.45	&&&	31.66	&&&	29.18	&&&	27.13	&&&	25.35	&&&	23.89	&&&	31.36	&&&	30.06	&&&	28.45	&&&	27.00	 &&&	25.67	&&&	24.48	&&&	28.10	&&&	27.49	&&&	26.76	&&&	25.98	&&&	 25.08	&&&	24.14	&&&	94.30	&&&	89.74	 &&&	84.88	&&&	79.50	&&&	73.97	&&&	68.67	&&&	93.62	&&&	90.31	&&&	86.36	&&&	82.00	&&&	77.57	&&&	73.39	 &&&	92.79	&&&	90.04	&&&	 86.51	&&&	82.62	&&&	78.46	&&&	74.49	\\
&&\bf{std}&&	35.87	&&&	31.72	&&&	29.13	&&&	27.12	&&&	25.36	&&&	23.86	&&&	35.58	&&&	31.46	&&&	28.97	&&&	27.16	 &&&	25.70	&&&	24.43	&&&	35.05	&&&	30.86	&&&	28.42	&&&	 26.65	&&&	25.23	&&&	24.09	&&&	94.55	&&&	89.63	 &&&	84.52	&&&	78.79	&&&	72.84	&&&	67.18	&&&	95.15	&&&	90.53	&&&	86.14	&&&	81.61	&&&	76.79	&&&	72.33	 &&&	95.06	&&&	 90.54	&&&	86.32	&&&	82.04	&&&	77.54	&&&	73.15	\\
&&\bf{max}&&	33.80	&&&	31.78	&&&	29.18	&&&	27.09	&&&	25.29	&&&	23.82	&&&	32.87	&&&	31.26	&&&	29.07	&&&	27.30	 &&&	25.77	&&&	24.46	&&&	31.00	&&&	29.94	&&&	28.34	&&&	 26.91	&&&	25.59	&&&	24.38	&&&	94.38	&&&	89.62	 &&&	84.53	&&&	78.83	&&&	72.97	&&&	67.37	&&&	94.16	&&&	90.42	&&&	86.16	&&&	81.57	&&&	76.92	&&&	72.59	 &&&	93.68	 &&&90.62	&&&	86.61	&&&	82.35	&&&	77.92	&&&	73.75	\\
&&\bf{heru}&&	35.89	&&&	31.81	&&&	29.25	&&&	27.23	&&&	25.42	&&&	23.92	&&&	35.79	&&&	31.72	&&&	29.31	&&&	 27.54	&&&	26.07	&&&	24.77	&&&	35.35	&&&	31.28	&&&	28.94	&&&	 27.25	&&&	25.88	&&&	24.71	&&&	94.51	&&&	 89.70	&&&	84.72	&&&	79.13	&&&	73.37	&&&	67.86	&&&	95.20	&&&	90.68	&&&	86.49	&&&	82.08	&&&	77.38	&&&	 72.98	&&&	95.33	 &&&\underline{90.98}	&&&\underline{87.05}&&&\underline{83.03}&&&\underline{78.69}&&&\underline{74.37}\\
&&\bf{stein}&&	35.67	&&&	31.77	&&&	29.18	&&&	27.09	&&&	25.29	&&&	23.82	&&&	35.51	&&&	31.52	&&&	29.15	&&&	 27.30	&&&	25.77	&&&	24.46	&&&	35.22	&&&	31.21	&&&	28.90	&&&	 26.96	&&&	25.59	&&&	24.38	&&&	94.45	&&&	 89.62	&&&	84.53	&&&	78.83	&&&	72.97	&&&	67.37	&&&	95.00	&&&	90.44	&&&	86.16	&&&	81.57	&&&	76.92	&&&	 72.59	&&&	95.21	&&&	90.78	 &&&	86.61	&&&	82.35	&&&	77.92	&&&	73.75	\\
&&\bf{ljs}&&	35.52	&&&	31.75	&&&	29.21	&&&	27.13	&&&	25.36	&&&	23.91	&&&	35.17	&&&	31.43	&&&	29.13	&&&	27.38	 &&&	25.90	&&&	24.63	&&&	35.16	&&&	31.08	&&&	28.77	&&&	 27.14	&&&	25.76	&&&	24.61	&&&	94.73	&&&	89.81	 &&&	84.55	&&&	78.62	&&&	73.09	&&&	68.29	&&&	94.59	&&&	89.95	&&&	85.58	&&&	80.41	&&&	75.31	&&&	70.59	 &&&	94.64	&&&	 89.62	&&&	85.35	&&&	80.79	&&&	75.84	&&&	71.08	\\
&&\bf{bss}&&\underline{36.15}&&&\underline{32.11}&&&\underline{29.80}&&&\underline{28.04}&&&\underline{26.49}&&&\underline{25.08}&&&\underline{35.89}&&&\underline{ 31.87}&&&\underline{29.47}&&&\underline{27.80}&&&\underline{26.41}&&&\underline{25.17}&&&\underline{35.44}&&&\underline{31.38}&&&\underline{29.05} &&&\underline{27.41}&&&\underline{26.03}&&&\underline{24.79}&&&\underline{94.92}&&&\underline{90.25}&&&\underline{86.74}&&&\underline{83.03}&&&\underline{ 78.80}&&&\underline{74.16}&&&\underline{95.25}&&&\underline{90.97}&&&\underline{86.79}&&&\underline{82.82}&&&\underline{79.18}&&&\underline{75.16}	&&&\underline{95.17}&&&	 90.61	&&&	86.14	&&&	 81.91	&&&	78.09	&&&	74.01	\\
\hline
\multirow{7}{*}{\begin{sideways}{\bf\textit{peppers}}\end{sideways}}&&\bf{zero}&&	32.49	&&&	29.69	&&&	27.49	&&&	25.69	 &&&	24.24	&&&	23.07	&&&	31.27	&&&	29.53	&&&	27.81	&&&	 26.30	&&&	25.01	&&&	23.91	&&&	30.81	&&&	29.24	 &&&	27.71	&&&	26.32	&&&	25.12	&&&	24.09	&&&	90.58	&&&	83.99	&&&	78.33	&&&	72.91	&&&	67.84	&&&	63.30	 &&&	89.93	&&&	 84.88	&&&	79.89	&&&	75.30	&&&	70.99	&&&	67.00	&&&	89.10	&&&	84.86	&&&	80.32	&&&	75.89	 &&&	71.81	&&&\underline{67.99}	\\
&&\bf{std}&&	33.72	&&&	29.84	&&&	27.45	&&&	25.62	&&&	24.15	&&&	22.97	&&&	33.46	&&&	29.91	&&&	27.73	&&&	26.07	 &&&	24.71	&&&	23.58	&&&	33.09	&&&	29.55	&&&	27.45	&&&	 25.85	&&&	24.53	&&&	23.47	&&&	91.20	&&&	84.07	 &&&	78.10	&&&	72.43	&&&	67.15	&&&	62.36	&&&	91.32	&&&	85.24	&&&	79.71	&&&	74.65	&&&	70.03	&&&	65.88	 &&&	90.68	&&&	 85.02	&&&	79.87	&&&	74.86	&&&	70.30	&&&	66.30	\\
&&\bf{max}&&	33.40	&&&	29.79	&&&	27.46	&&&	25.64	&&&	24.17	&&&	23.00	&&&	32.95	&&&	30.00	&&&	27.91	&&&	26.29	 &&&	24.94	&&&	23.80	&&&	32.72	&&&	29.95	&&&	27.97	&&&	 26.41	&&&	25.12	&&&	24.03	&&&	90.93	&&&	83.97	 &&&	78.11	&&&	72.49	&&&	67.24	&&&	62.49	&&&	91.11	&&&	85.05	&&&	79.71	&&&	74.89	&&&	70.40	&&&	66.31	 &&&	90.80	&&&	 85.35	&&&	80.30	&&&	75.58	&&&	71.25	&&&	67.31	\\
&&\bf{heru}&&	33.76	&&&	29.87	&&&	27.52	&&&	25.69	&&&	24.23	&&&	23.05	&&&	33.66	&&&	30.17	&&&	28.02	&&&	 26.38	&&&	25.01	&&&	23.89	&&&	33.36	&&&	30.01	&&&	28.02	&&&	 26.49	&&&	25.19	&&&	24.10	&&&	91.13	&&&	 84.07	&&&	78.23	&&&	72.66	&&&	67.48	&&&	62.79	&&&\underline{91.52}&&&\underline{85.44}&&&	80.00	&&&	75.18	&&&	70.72	 &&&	66.65	 &&&\underline{91.08}&&&\underline{85.72}&&&\underline{80.74}&&&	75.99	&&&	71.67	&&&	67.74	\\
&&\bf{stein}&&	33.45	&&&	29.79	&&&	27.46	&&&	25.64	&&&	24.17	&&&	23.00	&&&	33.35	&&&	30.00	&&&	27.91	&&&	 26.29	&&&	24.94	&&&	23.80	&&&	33.23	&&&	29.96	&&&	27.97	&&&	 26.41	&&&	25.12	&&&	24.03	&&&	90.93	&&&	 83.97	&&&	78.11	&&&	72.49	&&&	67.24	&&&	62.49	&&&	91.23	&&&	85.05	&&&	79.71	&&&	74.89	&&&	70.40	&&&	 66.31	&&&	91.02	&&&	85.35	 &&&	80.30	&&&	75.58	&&&	71.25	&&&	67.31	\\
&&\bf{ljs}&&	33.46	&&&	29.83	&&&	27.56	&&&	25.79	&&&	24.35	&&&	23.18	&&&	33.61	&&&	30.11	&&&	27.99	&&&	26.40	 &&&	25.04	&&&	23.91	&&&\underline{33.56}&&&\underline{30.10}	 &&&	28.02	&&&	26.45	&&&	25.20	&&&	24.11	&&&	91.02	&&&	84.23	 &&&	78.50	&&&	72.78	&&&	67.25	&&&	62.39	&&&	91.19	&&&	85.29	&&&	79.90	&&&	74.91	&&&	70.15	&&&	66.23	 &&&	 90.96	&&&	85.29	&&&	80.21	&&&	75.29	&&&	70.59	&&&	66.37	\\
&&\bf{bss}&&\underline{33.82}	&&&\underline{30.04}&&&\underline{27.85}&&&\underline{26.23}&&&\underline{24.93}&&&\underline{23.85}&&&\underline{33.69}&&&\underline{30.22} &&&\underline{28.11}&&&\underline{26.53}&&&\underline{25.25}&&&\underline{24.20}&&&	33.44	&&&	30.07	&&&\underline{28.06}	&&&\underline{26.55}&&&	 \underline{25.29}&&&\underline{24.25}&&&\underline{91.21}&&&\underline{84.71}&&&\underline{79.64}&&&\underline{75.07}&&&\underline{70.88}&&&\underline{66.90}&&& 91.26	&&&	85.38	 &&&\underline{80.42}&&&\underline{75.89}&&&\underline{71.71}&&&\underline{67.70}&&&	90.80	&&&	85.47	&&&	80.30	&&&\underline{76.07}	 &&&\underline{71.88}&&&\underline{67.99}\\
\hline
\multirow{7}{*}{\begin{sideways}{\bf\textit{house}}\end{sideways}}&&\bf{zero}&&	34.94	&&&	31.71	&&&	29.22	&&&	27.35	 &&&	25.92	&&&	24.84	&&&	34.44	&&&	32.18	&&&	29.94	&&&	28.10	 &&&	26.70	&&&	25.63	&&&	34.15	&&&	32.14	 &&&	30.12	&&&	28.35	&&&	26.94	&&&	25.84	&&&	88.89	&&&	83.08	&&&	77.62	&&&	72.21	&&&	67.19	&&&	62.60	 &&&	88.98	&&&	84.08	 &&&	79.30	&&&	74.71	&&&	70.56	&&&	66.90	&&&	88.72	&&&	84.43	&&&	80.11	&&&	75.59	 &&&	71.52	&&&	67.94	\\
&&\bf{std}&&	35.21	&&&	31.66	&&&	29.13	&&&	27.23	&&&	25.80	&&&	24.72	&&&	35.06	&&&	31.96	&&&	29.57	&&&	27.76	 &&&	26.39	&&&	25.33	&&&	34.81	&&&	31.77	&&&	29.44	&&&	 27.68	&&&	26.33	&&&	25.30	&&&	89.03	&&&	82.98	 &&&	77.26	&&&	71.53	&&&	66.12	&&&	61.18	&&&	89.10	&&&	83.91	&&&	78.78	&&&	73.90	&&&	69.56	&&&	65.91	 &&&	88.80	&&&	 84.07	&&&	79.20	&&&	74.33	&&&	70.04	&&&	66.45	\\
&&\bf{max}&&	35.19	&&&	31.68	&&&	29.15	&&&	27.26	&&&	25.83	&&&	24.75	&&&	35.21	&&&	32.28	&&&	29.86	&&&	27.99	 &&&	26.58	&&&	25.50	&&&	35.14	&&&	32.41	&&&	30.12	&&&	 28.28	&&&	26.82	&&&	25.71	&&&	88.96	&&&	82.98	 &&&	77.30	&&&	71.64	&&&	66.29	&&&	61.39	&&&	89.33	&&&	84.01	&&&	79.01	&&&	74.21	&&&	69.92	&&&	66.19	 &&&	89.49	 &&&\underline{84.51}&&&	79.86	&&&	75.14	&&&	70.90	&&&	67.24	\\
&&\bf{heru}&&	35.28	&&&	31.72	&&&	29.20	&&&	27.32	&&&	25.89	&&&	24.80	&&&	35.31	&&&	32.36	&&&	29.98	&&&	 28.09	&&&	26.68	&&&	25.59	&&&	35.10	&&&	32.40	&&&	30.24	&&&	 28.40	&&&	26.95	&&&	25.82	&&&	89.04	&&&	 83.05	&&&	77.43	&&&	71.85	&&&	66.64	&&&	61.83	&&&	89.41	&&&	84.17	&&&	79.22	&&&	74.48	&&&	70.21	&&&	 66.48	&&&	89.37	&&&	84.68	 &&&\underline{80.14}&&&	75.48	&&&	71.27	&&&	67.59	\\
&&\bf{stein}&&	35.19	&&&	31.68	&&&	29.15	&&&	27.26	&&&	25.83	&&&	24.75	&&&	35.26	&&&	32.28	&&&	29.86	&&&	 27.99	&&&	26.58	&&&	25.50	&&&	35.22	&&&	32.41	&&&	30.12	&&&	 28.28	&&&	26.82	&&&	25.71	&&&	88.96	&&&	 82.98	&&&	77.30	&&&	71.64	&&&	66.29	&&&	61.39	&&&	89.35	&&&	84.01	&&&	79.01	&&&	74.21	&&&	69.92	&&&	 66.19	 &&&\underline{89.52}&&&\underline{84.51}&&&	79.86	&&&	75.14	&&&	70.90	&&&	67.24	\\
&&\bf{ljs}&&	35.16	&&&	31.72	&&&	29.22	&&&	27.34	&&&	25.91	&&&	24.81	&&&	35.36	&&&	32.29	&&&	29.94	&&&	28.07	 &&&	26.65	&&&	25.54	&&&	35.39	&&&	32.36	&&&	30.12	&&&	 28.28	&&&	26.83	&&&	25.69	&&&	89.09	&&&	82.99	 &&&	77.31	&&&	71.52	&&&	66.27	&&&	61.76	&&&	89.44	&&&	83.79	&&&	78.70	&&&	73.59	&&&	68.60	&&&	64.01	 &&&	89.64	&&&	 84.08	&&&	79.15	&&&	74.13	&&&	69.27	&&&	64.62	\\
&&\bf{bss}&&\underline{35.37}&&&\underline{32.07}&&&\underline{29.86}&&&\underline{28.19}&&&\underline{26.86}&&&\underline{25.79}&&&\underline{35.40}&&&\underline{32.45}&&&\underline{30.25}	 &&&\underline{28.47}&&&\underline{27.14}&&&\underline{26.09}&&&\underline{35.43}&&&\underline{32.44}&&&\underline{30.31}&&&\underline{28.55}&&&\underline{27.14}&&&\underline{ 26.05}&&&\underline{89.45}&&&\underline{84.09}&&&\underline{79.86}&&&\underline{75.69}&&&\underline{71.53}&&&\underline{67.63}&&&\underline{89.59}&&&\underline{84.45} &&&\underline{80.21}&&&\underline{75.73}&&&\underline{71.47}&&&\underline{67.83}&&&\underline{89.52}&&&	84.32	&&&	80.10	&&&\underline{75.76}&&&\underline{ 71.53}&&&\underline{67.97}	\\
\hline
\multirow{7}{*}{\begin{sideways}{\bf\textit{lenna}}\end{sideways}}&&\bf{zero}&&	34.41	&&&	31.15	&&&	29.05	&&&	27.54	 &&&	26.38	&&&	25.49	&&&	33.91	&&&	31.47	&&&	29.57	&&&	28.14	 &&&	27.02	&&&	26.12	&&&	33.69	&&&	31.52	 &&&	29.72	&&&	28.34	&&&	27.23	&&&	26.34	&&&	89.31	&&&	82.44	&&&	77.26	&&&	72.75	&&&	68.49	&&&	64.49	 &&&	89.40	&&&	83.50	 &&&	78.70	&&&	74.81	&&&	71.51	&&&	68.69	&&&	89.13	&&&	84.05	&&&	79.33	&&&	75.48	 &&&	72.23	&&&\underline{69.43}\\
&&\bf{std}&&34.66	&&&	31.13	&&&	28.98	&&&	27.45	&&&	26.29	&&&	25.41	&&&	34.57	&&&	31.40	&&&	29.40	&&&	27.93	 &&&	26.81	&&&	25.92	&&&	34.35	&&&	31.30	&&&	29.35	&&&	 27.93	&&&	26.83	&&&	25.97	&&&	89.46	&&&	82.33	 &&&	76.97	&&&	72.09	&&&	67.46	&&&	63.08	&&&	89.82	&&&	83.44	&&&	78.31	&&&	74.24	&&&	70.83	&&&	68.02	 &&&	89.61	&&&	 83.77	&&&	78.71	&&&	74.60	&&&	71.25	&&&	68.46	\\
&&\bf{max}&&	34.63	&&&	31.13	&&&	29.00	&&&	27.47	&&&	26.31	&&&	25.42	&&&	34.71	&&&	31.55	&&&	29.53	&&&	28.06	 &&&	26.93	&&&	26.03	&&&\underline{34.76}&&&	31.77	&&&	 29.74	&&&	28.27	&&&	27.15	&&&	26.24	&&&	89.38	&&&	82.33	 &&&	77.01	&&&	72.20	&&&	67.62	&&&	63.30	&&&	89.92	&&&	83.45	&&&	78.46	&&&	74.44	&&&	71.05	&&&	68.20	 &&&	 90.08	&&&	84.14	&&&	79.14	&&&	75.12	&&&	71.78	&&&	68.93	\\
&&\bf{heru}&&	34.69	&&&	31.16	&&&	29.03	&&&	27.51	&&&	26.34	&&&	25.45	&&&	34.77	&&&	31.61	&&&	29.59	&&&	 28.13	&&&	27.00	&&&	26.09	&&&	34.64	&&&	31.74	&&&	29.81	&&&	 28.35	&&&	27.22	&&&	26.32	&&&	89.44	&&&	 82.39	&&&	77.11	&&&	72.41	&&&	67.94	&&&	63.74	&&&	90.01	&&&	83.62	&&&	78.63	&&&	74.62	&&&	71.24	&&&	 68.38	&&&	89.98	 &&&\underline{84.37}&&&	79.38	&&&	75.37	&&&	72.02	&&&	69.16	\\
&&\bf{stein}&&	34.63	&&&	31.13	&&&	29.00	&&&	27.47	&&&	26.31	&&&	25.42	&&&	34.72	&&&	31.55	&&&	29.53	&&&	 28.06	&&&	26.93	&&&	26.03	&&&\underline{34.76}	&&&	31.77	&&&	 29.74	&&&	28.27	&&&	27.15	&&&	26.24	&&&	89.38	 &&&	82.33	&&&	77.01	&&&	72.20	&&&	67.62	&&&	63.30	&&&	89.92	&&&	83.45	&&&	78.46	&&&	74.44	&&&	71.05	 &&&	68.20	&&&	 90.09	&&&	84.14	&&&	79.14	&&&	75.12	&&&	71.78	&&&	68.93	\\
&&\bf{ljs}&&		34.58	&&&	31.16	&&&	29.06	&&&	27.53	&&&	26.35	&&&	25.43	&&&	34.89	&&&	31.60	&&&	29.59	&&&	 28.11	&&&	26.94	&&&	25.99	&&&	35.00	&&&	31.78	&&&	29.76	&&&	 28.28	&&&	27.10	&&&	26.13	&&&	89.48	&&&	 82.63	&&&	76.96	&&&	71.65	&&&	66.69	&&&	62.17	&&&\underline{90.07}&&&	83.59	&&&	78.06	&&&	73.04	&&&	68.33	 &&&	64.01	 &&&\underline{90.22}&&&	84.03	&&&	78.45	&&&	73.34	&&&	68.65	&&&	64.32	\\
&&\bf{bss}&&\underline{34.78}	&&&\underline{31.43}	&&&\underline{29.52}	&&&\underline{28.16}&&&\underline{27.09}&&&\underline{26.22}&&&\underline{34.92}	&&&\underline{31.72} &&&\underline{29.80}&&&\underline{28.42}&&&\underline{27.35}&&&\underline{26.49}&&&\underline{34.76}&&&\underline{31.79}&&&\underline{29.92}&&&\underline{28.50}&&&\underline{27.41} &&&	 \underline{26.54}&&&\underline{89.72}&&&\underline{83.36}&&&\underline{78.88}&&&\underline{75.16}&&&\underline{71.78}&&&\underline{68.75}&&&	90.01	&&&\underline{84.05}&&&\underline{ 79.31}&&&\underline{75.24}&&&\underline{72.02}&&&\underline{69.08}&&&89.90	&&&	84.18	&&&\underline{79.51}&&&\underline{75.39}	&&&\underline{72.37}&&& 69.39	 \\
\hline
\multirow{7}{*}{\begin{sideways}{\bf\textit{barbara}}\end{sideways}}&&\bf{zero}&&	32.94	&&&	29.32	&&&	26.80	&&&	25.07	 &&&	23.87	&&&	23.02	&&&	32.30	&&&	29.86	&&&	27.74	&&&	 26.07	&&&	24.78	&&&	23.80	&&&	31.92	&&&	29.93	 &&&	28.09	&&&	26.58	&&&	25.32	&&&	24.30	&&&	91.63	&&&	82.83	&&&	73.96	&&&	66.25	&&&	60.30	&&&	56.05	 &&&	91.83	&&&	 85.55	&&&	78.31	&&&	71.30	&&&	65.12	&&&	60.12	&&&	91.54	&&&	86.57	&&&	80.40	&&&	74.06	 &&&	68.22	&&&	63.06	\\
&&\bf{std}&&	33.35	&&&	29.21	&&&	26.67	&&&	24.96	&&&	23.78	&&&	22.95	&&&	33.37	&&&	29.66	&&&	27.32	&&&	25.66	 &&&	24.43	&&&	23.52	&&&	33.19	&&&	29.62	&&&	27.44	&&&	 25.82	&&&	24.62	&&&	23.69	&&&	91.81	&&&	82.64	 &&&	73.55	&&&	65.73	&&&	59.77	&&&	55.42	&&&	92.32	&&&	85.02	&&&	77.27	&&&	69.95	&&&	63.74	&&&	58.88	 &&&	92.13	&&&	 85.44	&&&	78.39	&&&	71.56	&&&	65.48	&&&	60.51	\\
&&\bf{max}&&	33.34	&&&	29.28	&&&	26.73	&&&	25.00	&&&	23.81	&&&	22.97	&&&	33.39	&&&	30.04	&&&	27.68	&&&	25.96	 &&&	24.67	&&&	23.70	&&&	33.33	&&&	30.33	&&&	28.18	&&&	 26.50	&&&	25.20	&&&	24.19	&&&	91.76	&&&	82.67	 &&&	73.63	&&&	65.83	&&&	59.85	&&&	55.52	&&&	92.51	&&&	85.48	&&&	77.87	&&&	70.66	&&&	64.39	&&&	59.42	 &&&	92.66	&&&	 86.70	&&&	80.05	&&&	73.45	&&&	67.40	&&&	62.17	\\
&&\bf{heru}&&	33.40	&&&	29.33	&&&	26.78	&&&	25.05	&&&	23.85	&&&	23.00	&&&	33.58	&&&	30.07	&&&	27.77	&&&	 26.06	&&&	24.76	&&&	23.78	&&&	33.48	&&&	30.29	&&&	28.20	&&&	 26.59	&&&	25.31	&&&	24.28	&&&	91.83	&&&	 82.80	&&&	73.83	&&&	66.04	&&&	60.04	&&&	55.72	&&&	92.62	&&&	85.72	&&&	78.27	&&&	71.12	&&&	64.86	&&&	 59.82	&&&	92.62	&&&	86.93	 &&&	80.47	&&&	74.02	&&&	68.03	&&&	62.78	\\
&&\bf{stein}&&	33.34	&&&	29.28	&&&	26.73	&&&	25.00	&&&	23.81	&&&	22.97	&&&	33.40	&&&	30.04	&&&	27.68	&&&	 25.96	&&&	24.67	&&&	23.70	&&&	33.37	&&&	30.33	&&&	28.18	&&&	 26.50	&&&	25.20	&&&	24.19	&&&	91.76	&&&	 82.67	&&&	73.63	&&&	65.83	&&&	59.85	&&&	55.52	&&&	92.52	&&&	85.48	&&&	77.87	&&&	70.66	&&&	64.39	&&&	 59.42	&&&\underline{92.68}&&&	 86.70	&&&	80.05	&&&	73.45	&&&	67.40	&&&	62.17	\\
&&\bf{ljs}&&	33.25	&&&	29.32	&&&	26.80	&&&	25.07	&&&	23.87	&&&	23.02	&&&	33.68	&&&	30.04	&&&	27.77	&&&	26.07	 &&&	24.77	&&&	23.77	&&&\underline{33.79}&&&	30.31	&&&	 28.15	&&&	26.56	&&&	25.28	&&&	24.27	&&&	91.79	&&&	83.06	 &&&	74.55	&&&	67.20	&&&	61.06	&&&	56.04	&&&\underline{92.59}&&&	85.71	&&&	78.44	&&&	71.15	&&&	65.52	&&&	60.36	 &&&	92.67	&&&	86.60	&&&	80.38	&&&	74.06	&&&	68.18	&&&	62.80	\\
&&\bf{bss}&&\underline{33.47}&&&\underline{29.52}&&&\underline{27.15}&&&\underline{25.54}&&&\underline{24.38}&&&\underline{23.52}&&&\underline{33.62}&&&\underline{30.12}&&&\underline{27.87}	 &&&\underline{26.36}&&&\underline{25.03}&&&\underline{24.08}&&&	33.57	&&&\underline{30.37}&&&\underline{28.21}&&&\underline{26.67}&&&\underline{25.43}&&&\underline{24.44} &&&	 \underline{92.00}&&&\underline{84.14}&&&\underline{76.90}&&&\underline{70.26}&&&\underline{64.53}&&&\underline{59.79}&&&	92.48	&&&\underline{85.99}&&&\underline{79.34}&&&\underline{ 72.83}	 &&&\underline{67.33}&&&\underline{62.41}&&&	92.59	&&&\underline{86.97}&&&\underline{80.80}&&&\underline{74.87}&&&\underline{69.43}&&&\underline{64.46}\\
\hline
\end{tabular}
\end{table}
\end{landscape}

\begin{figure}[h]
  \scriptsize
  \begin{tabular}{@{}M@{}M@{}}
\includegraphics[width=7.5cm]{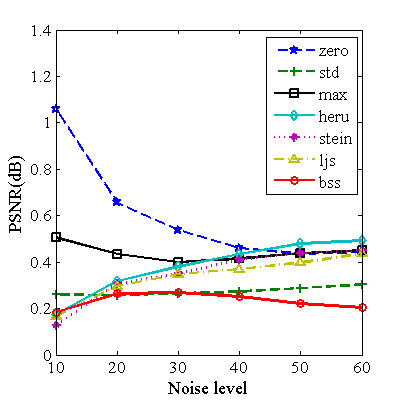}& \includegraphics[width=7.5cm]{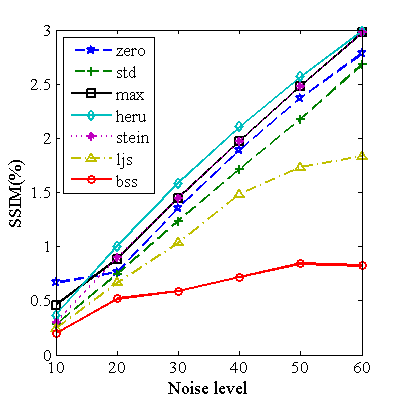}
  \end{tabular}
  \caption{The average of the standard PSNR and SSIM score deviation of tested NLM CPW/shrinkage solutions }
\end{figure}

\begin{figure}[!h]
  \scriptsize\centering
  \begin{tabular}{@{}r@{}M@{}m{.05cm}@{}M@{}r@{}M@{}m{.05cm}@{}M}
(a)&\includegraphics[width =3.9cm]{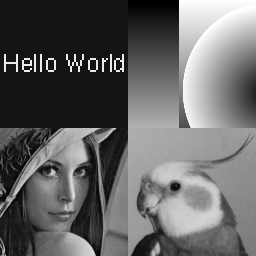} &&  & (b)&\includegraphics[width =3.9cm]{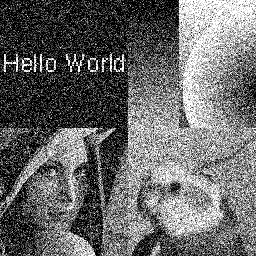} &&\includegraphics[width =3.9cm]{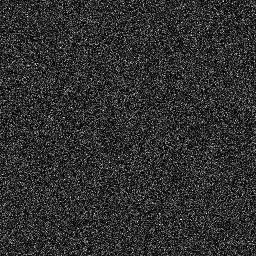}\\
(c)&\includegraphics[width =3.9cm]{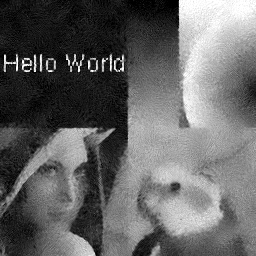} && \includegraphics[width =3.9cm]{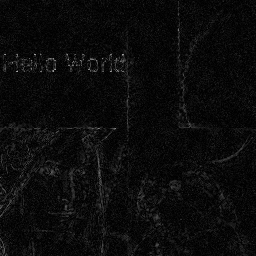} & (d)&\includegraphics[width =3.9cm]{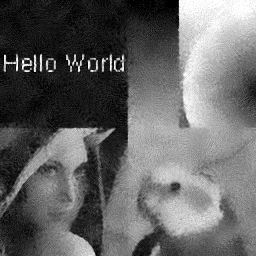} &&\includegraphics[width =3.9cm]{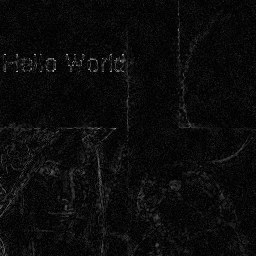}\\
(e)&\includegraphics[width =3.9cm]{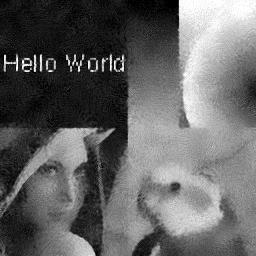} && \includegraphics[width =3.9cm]{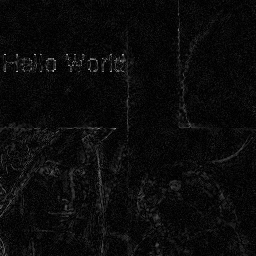} & (f)&\includegraphics[width =3.9cm]{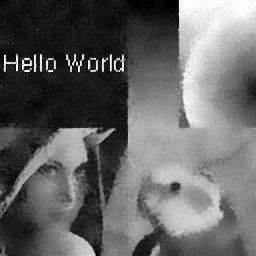} &&\includegraphics[width =3.9cm]{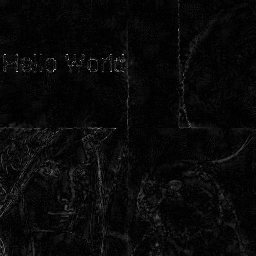}\\
  \end{tabular}
  \caption{Sample denoised images with corresponding absolute difference images from the clean image ($\sigma$=60, patch size is 3$\times$3). (a) original \textit{montage}, (b) observed noisy image and added noise, (c) to (f) best denoising results and corresponding method noise images by using method std, heur, ljs and bss, respectively. Corresponding PSNR/SSIM scores can be found in Table I.}
\end{figure}

\bibliographystyle{IEEEtran}
\bibliography{report}
\end{document}